\documentclass[final]{cvpr}
\usepackage{times}
\usepackage{epsfig}
\usepackage{graphicx}
\usepackage{amsmath}
\usepackage{amssymb}
\usepackage{bbm}
\usepackage{color}
\usepackage[title]{appendix}
\usepackage[pagebackref=true,breaklinks=true,colorlinks,bookmarks=false]{hyperref}

\begin{document}

\title{Breaking Shortcut: Exploring Fully Convolutional \\ Cycle-Consistency for Video Correspondence Learning}

\author{Yansong Tang\textsuperscript{1,}\thanks{\;indicates equal contribution, The work is done when Yansong Tang, Zhenyu Jiang and Zhenda Xie are interns at Microsoft Research Asia.},
    Zhenyu Jiang\textsuperscript{2,}$^*$,
    Zhenda Xie\textsuperscript{3,4,}$^*$,
    Yue Cao\textsuperscript{4},
 Zheng Zhang\textsuperscript{4},
 Philip H. S. Torr\textsuperscript{1},
 Han Hu\textsuperscript{4}\\ 
\textsuperscript{1} University of Oxford
\textsuperscript{2} University of Texas
\textsuperscript{3} Tsinghua University
\textsuperscript{4} Microsoft Research Asia\\
 {\tt \small \{yansong.tang,philip.torr\}@eng.ox.ac.uk; zhenyu@cs.utexas.edu;}\\
 {\tt \small xzd18@mails.tsinghua.edu.cn; \{yue.cao,zhez,hanhu\}@microsoft.com}}

\maketitle

\begin{abstract}
Previous cycle-consistency correspondence learning methods usually leverage image patches for training. In this paper, we present a fully convolutional method, which is simpler and more coherent to the inference process. While directly applying fully convolutional training results in model collapse, we study the underline reason behind this collapse phenomenon, indicating that the absolute positions of pixels provide a shortcut to easily accomplish cycle-consistence, which hinders the learning of meaningful visual representations. To break this absolute position shortcut, we propose to apply different crops for forward and backward frames, and adopt feature warping to establish correspondence between two crops of a same frame. The former technique enforces the corresponding pixels at forward and back tracks to have different absolute positions, and the latter effectively blocks the shortcuts going between forward and back tracks. In three label propagation benchmarks for pose tracking, face landmark tracking and video object segmentation, our method largely improves the results of vanilla fully convolutional cycle-consistency method, achieving very competitive performance compared with the self-supervised state-of-the-art approaches. Our trained model and code are available at \url{https://github.com/Steve-Tod/STFC3}.
\end{abstract}

\begin{figure}[t]
    \centering
    \includegraphics[width=1\linewidth]{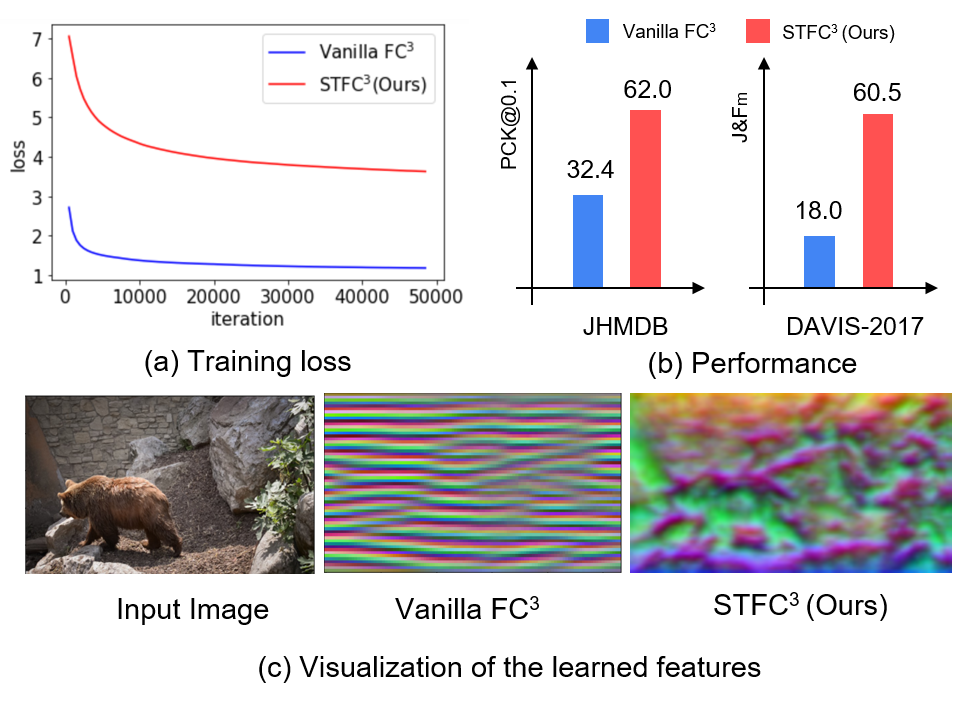}
    \caption{
    Comparison of vanilla fully convolutional cycle-consistency (FC$^3$) method and our proposed spatial transformation fully convolutional cycle-consistency (STFC$^3$) approach. 
    During training, the vanilla FC$^3$ converges very fast, as illustrated in the training loss curve (a). However, when evaluated on the pose tracking and video object segmentation tasks, this model shows inferior results (b). The visualization of its feature map in (c) further indicates that the model collapses into a shortcut solution. In this paper, we study the underline reason behind this phenomenon and introduce a new STFC$^3$ method to address this issue, which achieves significant improvements on the vanilla FC$^3$.
    }
    \vspace{-3mm}
    \label{fig:teaser}
\end{figure}


\begin{figure*}[t]
    \centering
    \vspace{0mm}
    \includegraphics[width=0.85\linewidth]{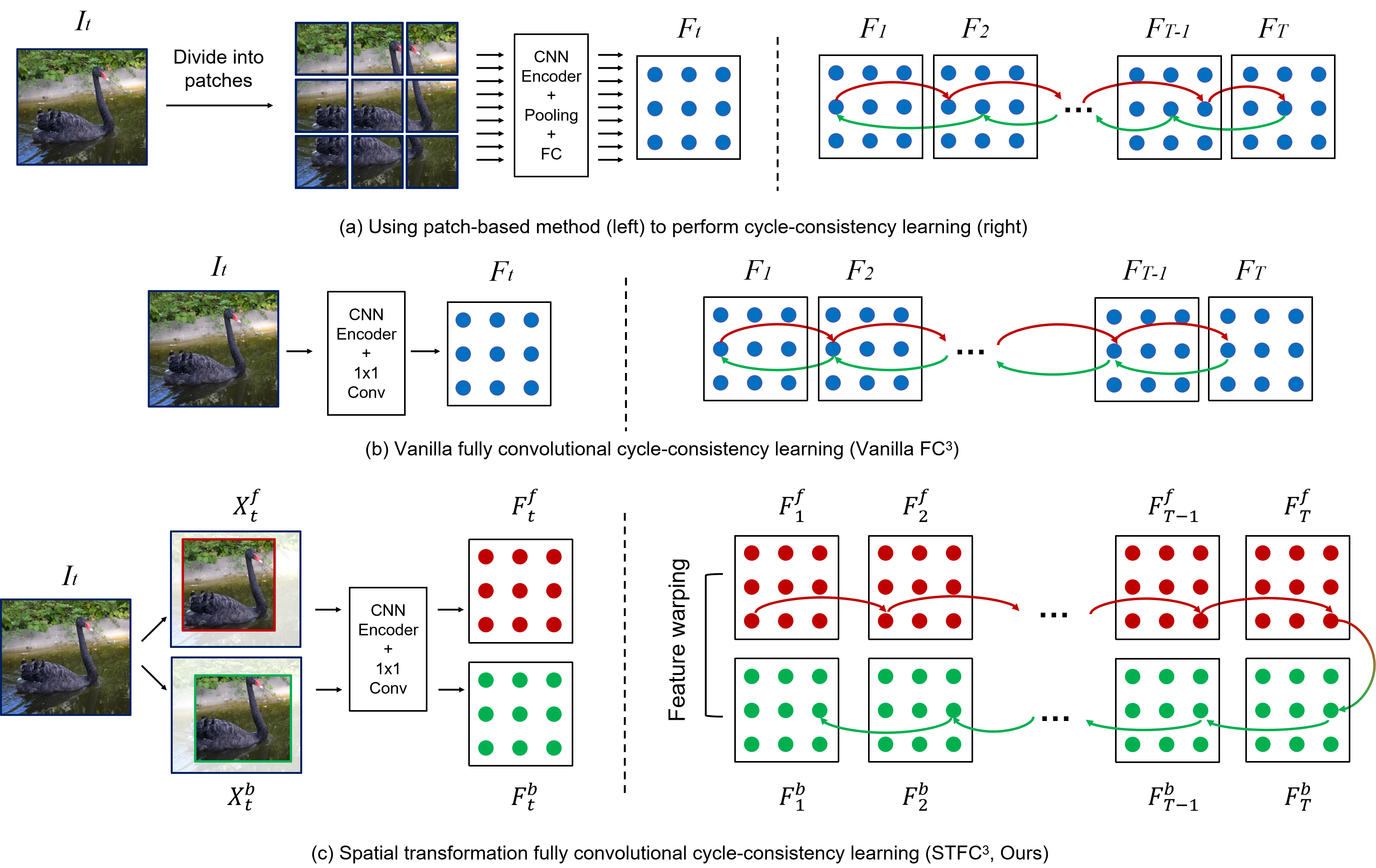}
    \caption{
    The state-of-the-art self-supervised video correspondence learning method CRW \cite{jabri2020walk} adopts a \textit{patch-based} method (a-left) to perform cycle-consistency learning (a-right), but extracts feature of the \textit{whole image} during testing.
    To bridge the gap between the training and testing, a more simple and effective way is to use a fully convolutional method during training (b).
    However, directly doing this causes the model to collapse into a shortcut solution as the CNN encodes absolute position information of each pixel. To address this issue, as shown in (c), we explore a spatial transformation fully convolutional cycle-consistency (STFC$^3$) method, which achieves significant improvements over three widely-used label propagation benchmarks.}
    \vspace{-3mm}    
    \label{fig:compare}
\end{figure*}

\section{Introduction}
Visual correspondence learning is a fundamental problem in computer vision.
In the past few years,
a variety of approaches~\cite{vondrick2018tracking, lai2019self, DBLP:conf/cvpr/LaiLX20, DBLP:conf/cvpr/LuWSTCH20} have been proposed to tackle this topic in a self-supervised learning paradigm,
by leveraging the inherent continuity between adjacent frames from unlabelled videos.
Among these works, a typical strategy is to take the cycle-consistency in time as free supervisory~\cite{wang2019learning, li2019joint, wang2019unsupervised}, and train an encoder tracking forward in time and backward to
the initial position to form a cycle.
Along this line, Allan \textit{et al.}~\cite{jabri2020walk} recently propose a method called ``contrastive random walk'' (CRW), which achieves state-of-the-art results
on various label propagation tasks. Specifically, CRW uses an encoder to learn representation of different \textit{patches} sampled from each frame (see Figure~\ref{fig:compare}(a)).
After training based on the cycle-consistency supervision, the optimized encoder is used to extract features of the \textit{whole image} for the downstream tasks.
Despite its elegant framework and promising performance, the inconsistency between its training and testing process motivates us to consider a natural question: \textit{can we apply a fully convolutional method rather than the patch-based one during the training process?}

Towards this direction, we attempt to directly execute cycle-consistency training based on a sequence of single feature maps extracted by a fully convolutional model (Figure \ref{fig:compare}(b)).
As shown in Figure~\ref{fig:teaser}(a), during training, this vanilla fully convolutional cycle-consistency (FC$^3$) converges very fast. But when evaluated on the pose tracking and video object segmentation tasks, this model shows inferior results (see Figure~\ref{fig:teaser}(b)). Based on this observation, we further visualize its feature map in Figure~\ref{fig:teaser}(c), which indicates that the model collapses into a shortcut solution during training\footnote{This phenomenon was also observed in the original CRW paper~\cite{jabri2020walk}}.
To better understand this collapse phenomenon, we make a throughout analysis and reveal that the underline reason behind is that the fully convolutional network encodes absolute position information of pixels~\cite{DBLP:conf/iclr/IslamJB20, jabri2020walk}, which provides a shortcut for the correspondence task and hinders the learning of meaningful visual representations.

To tackle this, we propose a \textbf{\textit{s}}\textit{patial} \textbf{\textit{t}}\textit{ransformation} \textbf{\textit{f}}\textit{ully} \textbf{\textit{c}}\textit{onvolutional} \textbf{\textit{c}}\textit{ycle}-\textbf{\textit{c}}\textit{onsistency} (STFC$^3$) approach, which applies different cropping operations when walking through forward and back tracks, and adopts feature warping other than probabilistic walking~\cite{jabri2020walk} to connect two crops of the same frame. The former technique lets the corresponding pixels at forward and back tracks have different absolute positions, and the latter effectively blocks the shortcuts going between forward and back tracks. 
Our STFC$^3$ effectively addresses the collapse issue. On three label propagation benchmarks, it largely improves the performance of vanilla FC$^3$ from 32.4$\%$ to 62.0$\%$ (PCK@0.1$\uparrow$) on J-HMDB~\cite{Jhuang:ICCV:2013} for pose tracking, from 56.7$\%$ to 18.8$\%$ (RMSE$\downarrow$) for on 300VW~\cite{DBLP:conf/iccvw/ShenZCKTP15} for face landmark tracking, and from 18.0$\%$ to 60.5$\%$ ($\mathcal{J}${\&}$\mathcal{F}_\textrm{m}$$\uparrow$) on DAVIS-2017~\cite{pont-tuset2017} for video object segmentation.
Compared with the self-supervised state-of-the-art methods, our STFC$^3$  achieves very competitive results on the unlabelled Kinetics dataset~\cite{DBLP:conf/cvpr/CarreiraZ17}.

\section{Related Works}
\noindent \textbf{Self-supervised Image Representation Learning:}
Self-supervised visual representation learning has attracted growing attention in recent years.
Currently, mainstream works~\cite{he2019moco, DBLP:conf/icml/ChenK0H20, BYOL, cao2020PIC, DBLP:conf/nips/CaronMMGBJ20} usually adopt an instance discrimination pretext task at image level, which regards each image as a single class and optimize the network to discriminate each other.
More recently, Xie \textit{et al.}~\cite{xie2020propagate} propose to perform unsupervised representation learning at pixel level, which effectively transfers performance to downstream tasks such as semantic segmentation and object detection.
While it is a natural way to borrow the merit of recent progress~\cite{xie2020propagate} from image domain to video domain, the trivial solution caused by the position encoding becomes a main obstacle for correspondence learning. In this paper, we make a throughout analysis on this point and propose to break the shortcut with a spatial transformation strategy.

\noindent \textbf{Self-supervised Video Representation Learning:}
In the past few years, there are growing numbers of works dedicated to self-supervision video representation learning for various downstream tasks, such as action recognition~\cite{DBLP:conf/nips/HanXZ20, DBLP:conf/eccv/HanXZ20,cvrl}, video retrieval~\cite{DBLP:conf/cvpr/MiechASLSZ20}, video caption~\cite{DBLP:conf/iccv/SunMV0S19, DBLP:conf/cvpr/ZhuY20a} and many others. 
In this paper, we focus on the downstream task of label propagation.
According to the type of self-supervision used for training,
we divide existing approaches into colorization-based, cycle-consistency-based and frame-level similarity learning methods.
For the first category, Vondrick \textit{et al.}~\cite{vondrick2018tracking} and Lai \textit{et al.}~\cite{DBLP:conf/bmvc/LaiX19} utilize the pretext task to reconstruct a target frame from a reference frame.
More recently, Lai \textit{et al.}~\cite{DBLP:conf/cvpr/LaiLX20} introduce a memory component that allows the network to capture more contextual information of different frames to enhance the existing methods.
For the second category, Wang \textit{et al.}~\cite{wang2019learning} and \cite{wang2019unsupervised} proposed to use cycle-consistency in time as free supervision, to train an encoder tracking forward in time and backward to form a cycle. More recently, Allan \textit{et al.}~\cite{jabri2020walk} propose a probabilistic framework that largely improves the performance on various tasks. We also follow this line but explore a new fully convolutional method for cycle-consistency learning. Concurrently, Xu \textit{et al.}~\cite{VFS} and Mathilde \textit{et al.}~\cite{dino} show that using frame-level similarity could also learn the correspondence in an implicit way, which are complementary methods with ours.

\noindent \textbf{Position Information Encoded by Neural Network:}
Recent studies~\cite{DBLP:conf/iclr/IslamJB20,DBLP:conf/cvpr/KayhanG20} have shown that convolutional neural networks can encode the position information of the input image, mainly caused by the padding operation. 
There also have been a series of works explicitly leverage the position information for devising different architectures (\textit{e.g.,} Transformer network~\cite{DBLP:conf/nips/VaswaniSPUJGKP17}, local relation network~\cite{DBLP:conf/iccv/HuZXL19} and generative adversarial network~\cite{DBLP:journals/corr/abs-2012-05217}) or enhancing the performance for various computer vision tasks, such as instance segmentation~\cite{DBLP:conf/cvpr/NevenBPG19}, refer expression segmentation~\cite{DBLP:conf/iccv/LiuLSYLY17}, \textit{etc}.
On the contrary, for cycle-consistency corresponding learning, we find the position information causes the neural network to collapse into a shortcut solution, and aim to eliminate this effect during the self-supervised learning stage.

\section{Proposed Approach}

\subsection{Revisiting the CRW}
During training, CRW~\cite{jabri2020walk} first unfolds a sequence of video frames into different patches $\{\{P_{t}^i\}_{t=1}^T\}_{i=1}^N$, as illustrated in Figure \ref{fig:compare} (a-left).
The transition matrix between two adjacent frames is defined as:
\begin{align}
     A_t^{t+1}(i,j) = \frac{\exp({{<\phi(P_t^i),  \phi(P_{t+1}^j)>}/{\tau}})}{\sum_{l} \exp({<\phi(P_t^i),  \phi(P_{t+1}^l)>}/{\tau})},
    \label{eq:affinity}
\end{align}
where $\phi$ denotes a ResNet18 encoder followed by a pooling and an fully-connected layer, $<, >$ represents inner product and $\tau$ is a temperature factor.
Then a long-range walking for timestep 1 to timestep $T$ could be calculated as $\bar{A}_1^{T}= \prod_{i=1}^{T-1} {A_{i}^{i+1}}$.
To leverage temporal cycle-consistency as free supervision, CRW concatenates the sequence with a temporally reversed version of itself.
After walking through this palindrome, the expected terminal of each query node is its initial position, which formulates the loss function as:
\begin{align}
  \mathcal{L}^{k}_{cyc} &= \mathcal{L}_{CE}(\bar{A}_1^{T} \bar{A}_{T}^1, I).
  \label{eq:losscyc_0}
\end{align}
During testing, the trained model is used to extract the feature map of the whole video frame.
Then the similarity between the feature nodes from adjacent frames is calculated, which guides the ground-truth label given in the first frame to propagate to the later frames progressively.

\subsection{Vanilla FC$^3$ and its Shortcut}
In order to eliminate the inconsistency between the training and testing process of CRW, 
it is straightforward to explore a fully convolutional cycle-consistency correspondence learning during training.
Towards this direction, we first make an attempt as shown in Figure \ref{fig:compare} (b).
Specifically, given a sequence of video frames $\{I_1, I_2, ..., I_T\}$ with the size of $T \times H \times W \times 3$, we use a ResNet-18 encoder followed by a fully-connected layer, denoted as $\Psi$, to extract feature of each image directly, rather than first dividing them into patches. 
We obtain a set of feature map $\{\phi(I_1), \phi(I_2), ..., \phi(I_T)\}$ with the size of $T \times H' \times W' \times C$.
As previously, we consider them as a space-time graph, where $H' \times W'$, $T$ and $C$ corresponds to the spatial dimension, temporal dimension, and channel dimension, respectively. 
Different from Eqn. (\ref{eq:affinity}), the transition matrix in this case is defined as:
\begin{align}
     A_t^{t+1}(i,j) = \frac{\exp({{<\phi^i(I_t),  \phi^j(I_{t+1})>}/{\tau}})}{\sum_{l} \exp({<\phi^i(I_t),  \phi^l(I_{t+1})>}/{\tau})}.
    \label{eq:affinity1}
\end{align}

Here $\phi^i(I_t)$ denotes the $i$-th node in the feature map $\phi(I_t)$. Then, we could perform the cycle-consistency learning similar to Eqn. (\ref{eq:losscyc_0}). However, after doing this, we found the performance drops significantly (See Figure~\ref{fig:teaser}). As indicated by previous work~\cite{jabri2020walk}, this is probably because the convolutional architecture encodes the position information of the input image, thus the network can perform correspondence learning based on the position information of each node, rather than its appearance information. 

\begin{figure}[t]
    \centering
    \includegraphics[width=1\linewidth]{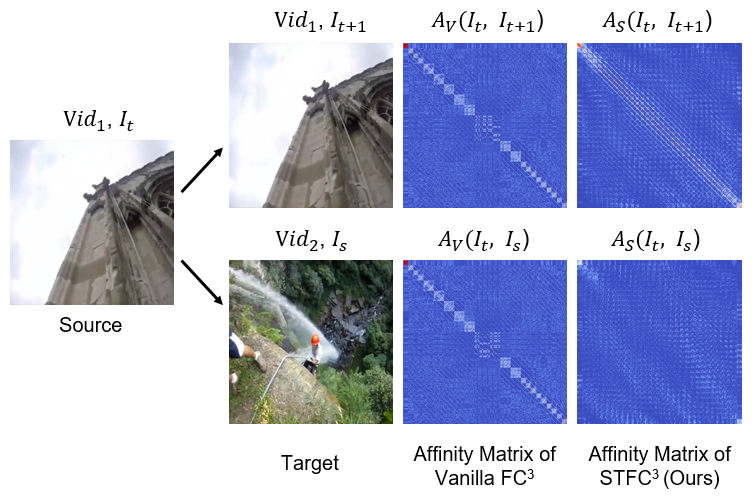}
    \caption{
    Visualization of affinity matrices between features given by vanilla FC$^3$ and our STFC$^3$. These matrices are computed following Eqn. (\ref{eq:affinity1}). A brighter pixel indicates a higher similarity between feature points. Vanilla FC$^3$ gives very similar affinity matrices between a pair of alike frames ($I_t$ and $I_{t+1}$) and a pair of very different frames ($I_t$ and $I_{s}$), while STFC$^3$ shows plausible correspondence in the affinity matrices between the two pairs. This indicates vanilla FC$^3$ learns the correspondence based on the position rather than visual appearance.
    }
    \label{fig:vis_aff}
\end{figure}

To make a deeper understanding of this, we present visualization results in Figure \ref{fig:vis_aff}. In the first row, we show the affinity matrix between two neighboring frames ($I_t$ and $I_{t+1}$) from the same video. Both affinity matrices from vanilla FC$^3$ and STFC$^3$ (will be detailed later) show a pattern of large diagonal items. This is plausible because the matching target features of source features are in the neighboring position. In the bottom row, we show the affinity matrix between two very different frames. In this case, there is no strong pattern in the affinity matrix of STFC$^3$ since there are no matching points between the source and target frames. However, vanilla FC$^3$ still gives an affinity matrix with large diagonal items. This indicates vanilla FC$^3$ learns correspondence by encoding positional information rather than visual information in the feature map. The reason behind this is that vanilla FC$^3$ learns a shortcut when trained with cycle consistency loss. The cycle consistency loss enforces a near identity transition matrix of a cycle. The transition matrix of a cycle is composed of affinity matrices of adjacent frames. So vanilla FC$^3$ can encode positional information in the features, which induce near identity affinity matrices and transition matrix of the cycle. With this shortcut, vanilla FC$^3$ can fit the training target while ignoring the visual correspondence.

In order to break this shortcut, we explore a variety of attempts. We introduce our spatial transform FC$^3$ as below, while other strategies could be found in the experiment part.

\subsection{Spatial Transformation FC$^3$}
Figure \ref{fig:compare} (c) illustrates the pipeline of our method. To break the shortcut learned by vanilla FC$^3$, our key idea is to perform different spatial transformation operations on the same frame to get different copies of a raw frame. For a certain pixel in the raw frame, the positions of the corresponding pixels in these copies are different since they undergo different operations. The position information can no longer guide the network to learn correspondence. In this way, the network is enforced to learn meaningful representations based on the visual appearance rather than the position information. In this subsection, we will introduce the spatial transformation operations, as well as how we align the nodes to form a cycle and calculate a new loss after the spatial transformation.

\noindent \textbf{Spatial Transformation:} As previous, suppose we have a sequence of video frames $\{I_1, I_2, ..., I_T\}$, we first perform different spatial transformation operations to the forward and backward frames respectively: 
\begin{align}\label{con:trans}
    X_t^{f} = \Omega_{\theta_f}(I_t), \; X_t^{b} = \Omega_{\theta_b}(I_t), \; t=1, 2, ..., T.
\end{align}

For different transformations $\Omega_{\theta_f}$ and $\Omega_{\theta_b}$, we execute random resized crop random horizontal flip on the original frames, and resize them to the same size of $[H_1, W_1, 3]$. This kind of spatial transformation could actually be determined by an affine matrix $B_{\theta}$ with hyper-parameter $\theta$~\cite{DBLP:conf/nips/JaderbergSZK15},
which transforms the source coordinates to the target ones. We leverage $B_{\theta}$ to warp feature, which will be detailed later. 
Similar to vanilla FC$^3$, we adopt a feature extractor $\psi$ containing a ResNet-18 encoder and an 1 $\times$ 1 convolutional layer to extract features of these augmented frames:
\begin{align}
    F_t^{f} = \psi(X_t^{f}), \;  F_t^{b} = \psi(X_{t}^{b}), \; t = 1, 2, ... ,T.
\end{align}

Through the transformation Eqn.(\ref{con:trans}), the same region of the raw input $I_i$ corresponds different nodes of forward features $F_i^{f}$ and backward features $F_i^{b}$. 

\noindent \textbf{Walking through a Palindrome: }
After spatial transformation, we concatenate the obtained feature to form a new sequence:
\begin{align}
    S = [F_1^{f},  \cdots, F_{T}^{f}, F_{T}^{b}, \cdots, F_{1}^{b}].
\end{align}

The walk between two adjacent frames is slightly different as previous, which could be formulated as:
\begin{align}
     \{A^{S}_{t}\}_{i,j} = \frac{\exp({{<S^{i}_t, S^{j}_{t+1}>}/{\tau}})}{\sum_{l} \exp({<S^{i}_t, S^{l}_{t+1}>}/{\tau})}.
\end{align}

To form a palindrome, a node first walks from timestep 1 to $T$ in the forward track, then moves to timestep $T$ in the backward track, and finally returns to timestep 1. The whole process could be written as: 
\begin{align}
A_{cyc}^S = \prod_{t=1}^{2T-1} A^{S}_{t}.
\end{align}

\noindent \textbf{Feature Warping:}
With spatial transformation, the identity mapping between the nodes of the beginning and the ending of a palindrome is not preserved anymore, because the forward frame $I^{f}_t$ and the corresponding backward frame $I^{b}_{t}$ undergo different spatial transformation and are not aligned anymore. To this end, we introduce a feature warping strategy to find the mapping between the start nodes $F_1^{f}$ and end nodes $F_1^{b}$ of a palindrome.

When transforming the forward and backward frames, we collect the corresponding affine matrices $B_{\theta_f}, B_{\theta_b}$
Then we compute the affine matrix that can align the forward frame with the backward frame by $B_{fb} = B_{\theta_b} B_{\theta_f}^{-1}$, 
and apply it to the first feature map $F_1^{f}$ in the forward track as: 
\begin{align}
F_1^{fb} = \mathcal{T}_{B_{fb}}(F_1^{f}),
\label{eq:wrapping}
\end{align}
where $\mathcal{T}_B(F)$ means applying transformation with affine matrix $B$ to a 2D feature $F$. After this warping, the nodes of $F_1^{fb}$ become aligned with the backward feature map $F_1^{b}$. Notice here, after the spatial feature warping, the position-feature mapping is changed. In this way, we can keep the identity mapping between the start and end nodes while factoring out the position information. The sequence of features after warping is:
\begin{align}
    S_w = [F_1^{fb},  \cdots, F_{T}^{f}, F_{T}^{b}, \cdots, F_{1}^{b}].
\end{align}

\begin{figure}[t]
    \centering
    \includegraphics[width=0.8\linewidth]{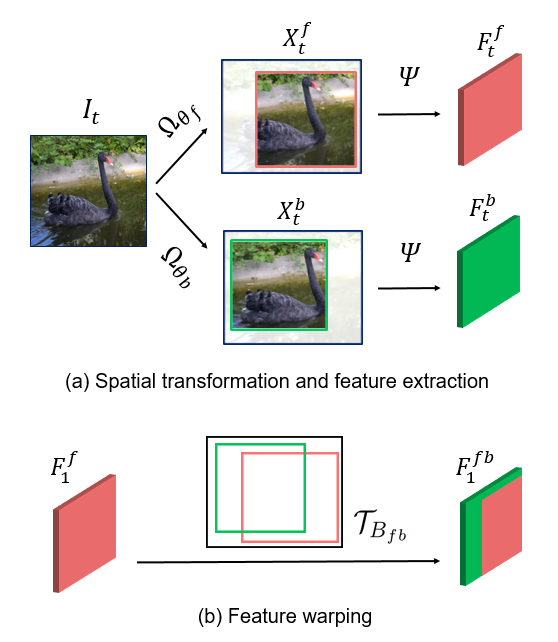}
    \vspace{1mm}
    \caption{(a) We employ different spatial transformation $\Omega_{\theta_f}$, $\Omega_{\theta_b}$ to the frames in the forward track and backward track, so that the same content in the original input $I_t$ corresponds to different positions in the extracted feature maps $F_t^{f}$ and $F_t^{b}$.
    (b) Illustration of the feature warping. According to~\cite{DBLP:conf/nips/JaderbergSZK15}, the spatial transformation $\Omega_{\theta_f}$ and $\Omega_{\theta_b}$ corresponds to two affine matrices $B_{\theta_f}$ and $B_{\theta_b}$. We use $B_{fb} = B_{\theta_b} B_{\theta_f}^{-1}$ as the warping matrix, and perform the corresponded transformation $\mathcal{T}_{B_{fb}}$ on $F_1^{f}$. With the obtained feature $F_1^{fb}$ and other features, we form a new cycle $S_w = [F_1^{fb},  \cdots, F_{T}^{f}, F_{T}^{b}, \cdots, F_{1}^{b}]$ for self-supervised learning.
    }
    \label{fig:method}
\end{figure}

\noindent \textbf{Mask Loss:}
The random crop for forward and backward frames can be different. When the crop region of backward frames is not contained in the counterpart of forward frames, the warped forward frame $F_1^{fb}$ would have zero margins. And the nodes in these zero margins do not correspond to any of the end nodes. To tackle this problem, we compute a mask by applying the same affine transformation to an all-one 2D tensor:
\begin{align}
M = \mathcal{T}_{B_{fb}}(\mathbbm{1}).
\end{align}
Here $\mathbbm{1}$ is an all-one 2D tensor with the same spatial size as $F_1^{f}$. Then we binarize $M$ using 0.5 as threshold to get binary mask $M_b$. We mask out the losses of all walks that start from a node in zero margin, which are marked as 0 in $M_b$. Finally, we optimize the total loss as follows:
\begin{align}
      \mathcal{L}^{S_w}_{cyc} &= \frac{1}{H_2 W_2} \sum_{i=1}^{H_2 W_2} M_b (i) \mathcal{L}_{CE}(A_{cyc}^{S_w} (i), i) \\
      &= -\frac{1}{H_2 W_2} \sum_{i=1}^{H_2 W_2} M_b (i) \log (A_{cyc}^{S_w} (i, i)).
  \label{eq:losscyc}
\end{align}

Here $H_2$ and $W_2$ are the height and width of the feature map. Given a video sequence of length T, we train our model on the multiple cycles of the sequence $S_w^1, ..., S_w^{T-1}$, where
$
      S_w^k = [F_k^{fb},  \cdots, F_{T}^{f}, F_{T}^{b}, \cdots, F_{k}^{b}].
$
And the final loss involves cycles of multiple lengths during training
$
    \mathcal{L}_{cyc} = \sum_{k=1}^{T-1} \mathcal{L}^{S_w^{k}}_{cyc}.
$

\begin{table*}[t]
	\centering
	\caption{Experiment results on different methods to avoid the shortcut solution. Our method achieves significant improvement over the vanilla FC$^3$ method on three label propagation tasks. Experiments are conducted on the J-HMDB~\cite{Jhuang:ICCV:2013}, 300VW~\cite{DBLP:conf/iccvw/ShenZCKTP15} and DAVIS-2017~\cite{pont-tuset2017} for pose tracking, face landmark tracking and video object segmentation respectively.}
	\label{tab:padding}
		\vspace{2mm}
	\begin{tabular}{l|cc|c|ccc}
			\hline
Task  & \multicolumn{2}{c|}{Pose Tracking} & Face Landmark Tracking  &\multicolumn{3}{|c}{Video Object Segmentation}
\\			
			\hline
{Metric}  & PCK@0.1 $\uparrow$ & PCK@0.2 $\uparrow$ & RMSE $\downarrow$ & $\mathcal{J}${\&}$\mathcal{F}_\textrm{m} \uparrow$ & $\mathcal{J}_\textrm{m} \uparrow$  & $\mathcal{F}_\textrm{m} \uparrow$  \\
			\hline
Vanilla FC$^3$ (Zero Padding) & 32.4 & 48.4 & 56.7 & 18.0 & 15.7  & 20.2 
\\
FC$^3$ (Replicate Padding) & 49.7 & 67.6 & 28.2 & 31.5 & 29.8  & 33.3 
\\
FC$^3$ (Reflect Padding) & 45.9 & 63.7 & 26.9 & 28.8 & 26.3  & 32.0
\\
FC$^3$ (No Padding)  & 35.1 & 52.7 & 50.2& 38.8 & 35.6 & 41.9 
\\
			\hline
STFC$^3$ (Ours)   & \textbf{62.0} & \textbf{80.5} & \textbf{18.8} & \textbf{60.5} & \textbf{58.0}  & \textbf{63.1} \\
\hline
	\end{tabular}
\end{table*}

\begin{table*}[t]
	\centering
\caption{Evaluation of the pose tracking task with the J-HMDB benchmarks. SM, D and I represent using Sintel Movie~\cite{LevyR10Sintel}, DAVIS2017 and ImageNet as the training data respectively.}
\vspace{2mm}
\label{tab:pose_tracking}
	\begin{tabular}{lccccc}
			\hline
 Method & Supervised? & Training Data & Backbone & PCK@0.1 & PCK@0.2   \\
 			\hline
Thin-Slicing Network~\cite{DBLP:conf/cvpr/0006WGH17} & $\surd$ & J-HMDB+I & Self-Designed & 68.7 & 92.1 \\
PAAP~\cite{DBLP:conf/fgr/IqbalGG17} & $\surd$ & J-HMDB+I & VGG-16 & 51.6 & 73.8 \\
ResNet-18 \cite{He2015a}& $\surd$ & ImageNet & ResNet18 & 59.0 & 80.6 \\ 
\hline
MoCo~\cite{he2019moco} &$\times$& ImageNet  & ResNet18 & 58.1 & 75.6 \\ 
VINCE~\cite{gordon2020watching}  &$\times$& Kinetics & ResNet18 & 58.4 & 75.7 \\ 
\hline
Identity &$\times$& - &  - & 43.1 & 64.5   \\ 
ColorPointer~\cite{vondrick2018tracking} &$\times$& Kinetics &  ResNet18 & 45.2 & 69.6   \\ 
TimeCycle~\cite{wang2019learning} &$\times$& VLOG &  ResNet18 & 57.3 & 78.1   \\ 
mgPFF~\cite{kong2019mgPFF} &$\times$& SM+D+J-HMDB &  ResNet18 & 58.4 & 78.1   \\ 
UVC \cite{li2019joint} &$\times$& Kinetics & ResNet18 & 58.6 &  79.6 \\ 
CRW~\cite{jabri2020walk}  &$\times$& Kinetics & ResNet18 & 58.8 & 80.2 \\ 
VFS~\cite{VFS} & $\times$ & Kinetics & ResNet18 & 60.5 & 79.5 \\ 
STFC$^3$ (Ours)  &$\times$& Kinetics & ResNet18 & \textbf{62.0} & \textbf{80.5} \\
\hline
	\end{tabular}
\end{table*}

\section{Experiments}
\subsection{Datasets and Experiment Settings}
We train our model from scratch in the Kinetics dataset~\cite{DBLP:conf/cvpr/CarreiraZ17}, without using any human annotation. 
The evaluation is performed on three label propagation tasks. Given the ground-truth labels of the first frames, the goal is to predict the label of the later frames.
We introduce the datasets used for different tasks as follows.

\noindent \textbf{J-HMDB~\cite{Jhuang:ICCV:2013}:}
We conduct experiments in J-HMDB dataset for pose tracking, following the protocol adopted in ~\cite{li2019joint, jabri2020walk}. There are 268 videos for testing, where each video contains about 40 frames with 15 keypoints of body.
We use frames with 640$\times$640 resolution as the input of the model.
We report the performance of different methods with the PCK@$\alpha$ evaluation metric, which denotes the Probability of Correct
Keypoint at a threshold of $\alpha$.

\noindent \textbf{300VW~\cite{DBLP:conf/iccvw/ShenZCKTP15}:}
The 300VW is a widely used video dataset for face landmark tracking. 
We evaluate our model on split-1 which contains 31 videos.
This task is more difficult than the pose tracking task, as there are thousands of frames in each video and each video frame contains 68 face landmarks.
We follow~\cite{DBLP:conf/aaai/ZhuLYS20} to use the cropped face as input and resized to the size of 320 $\times$ 320.
We employ the point-to-point Root-Mean-Square Error (RMSE) defined in~\cite{DBLP:conf/iccvw/ShenZCKTP15} as the evaluation metric.

\noindent \textbf{DAVIS-2017~\cite{pont-tuset2017}:}
We study the video object segmentation task on the validation set of DAVIS-2017, which contains 30 Youtube videos.
We use 720p resolution images as input, and evaluate our method based on the \textit{m} (mean) of Jacaard index $\mathcal{J}$ and contour-based accuracy $\mathcal{F}$ metrics~\cite{pont-tuset2017}.

\subsection{Implementation Details}
During the training process, we adopted ResNet-18 encoder followed by 2 fully-connected (fc) layers to extract the feature map of the input video frames. The dimensions of the fc layers were 2048 and 512. We used the Adam optimizer with the initial learning rate of 1e-4 to train our model. The batch size, temperature factor $\tau$, and the cycle-length was set to be 64, 0.05 and 4, respectively. A random resized crop with size $~U(0.08, 1)$ and a random horizontal flip with probability 0.5 were used as the spatial transformation. We also applied color augmentations in \cite{BYOL} to all the frames. We followed ~\cite{jabri2020walk} to modify the stride of the layer3 and layer4 from 2 to be 1, resulting in $32 \times 32$ nodes for random walking.
We trained our model with 8 V100 GPUs and found it converged at the first epoch on the Kinetics dataset, The training process only costed about 4.5 hours, largely saving the time compared with previous self-supervised learning works~\cite{jabri2020walk,li2019joint}.

During the testing process, we removed the fc layers and the 4-th ResNet18 layer of the trained model to extract features, obtaining a set of source nodes of the first frame and the target nodes of the later frame. Then we use Eqn. (\ref{eq:affinity1}) to calculate the similarity of each node, and consider the top-\textit{k} transitions to perform label propagation~\cite{jabri2020walk, wang2019learning, li2019joint}. We also adopted the temporal context frames $m$ and restricted the label radius $r$~\cite{jabri2020walk} to achieve better performance. More details of the hyper-parameters we used could be found in the supplementary material.

\subsection{Study of Different Methods to Avoid the Shortcut Solution}
Through the visualization results in Figure \ref{fig:vis_aff}, we find that the position encoded by the CNN model is the main reason to cause the shortcut solution. According to the previous work~\cite{DBLP:conf/iclr/IslamJB20}, the position information is mainly introduced by the zero paddings. So a straightforward way to avoid the shortcut solution is to use different padding methods (e.g., replicate padding and reflect padding). Another strategy is to give up padding operation and crop the feature boundary as the network goes in deeper~\cite{jabri2020walk}.
We first evaluate these methods and present the results in Table \ref{tab:padding}.
We observe that these strategies can improve the baseline accuracy and alleviate the shortcut effect, but the performance is still not promising.  
This is because changing padding methods do not eliminate the source of the position information, although different padding methods make it harder to learn to encode position information.
In comparison, our STFC$^3$ achieves the improvements of 29.6 PCK@0.1 on J-HMDB for pose tracking, -37.4 RMSE($\downarrow$) on 300VW for face landmark tracking and 42.5 $\mathcal{J}${\&}$\mathcal{F}_\textrm{m}$ on DAVIS-2017 for video object segmentation.
This strongly demonstrates the effectiveness of our method to address the shortcut issue.

\subsection{Comparison with the State-of-the-art}
\begin{figure}[t]
    \centering
    \includegraphics[width=1\linewidth]{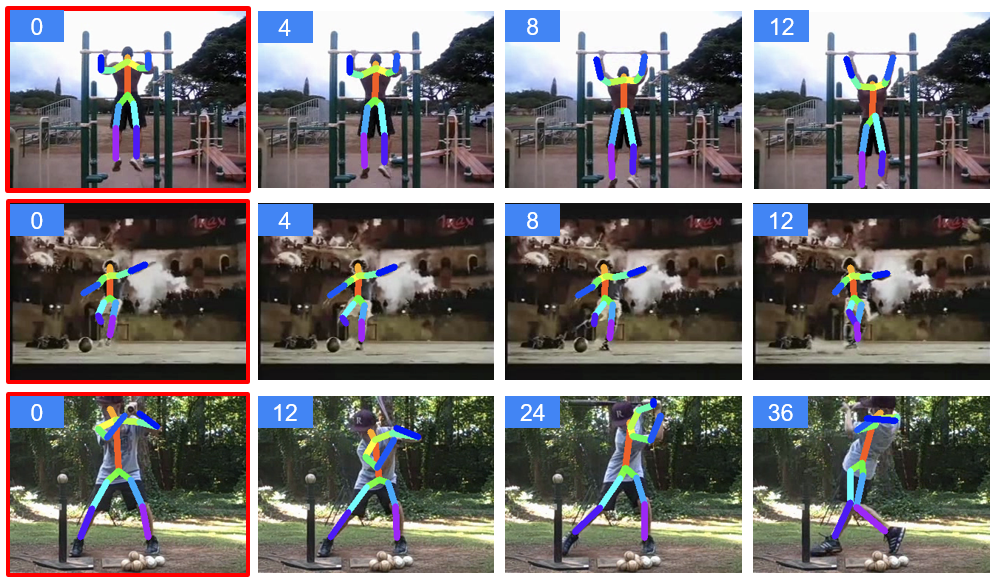}
    \caption{Visualization of the pose tracking results in the J-HMDB dataset~\cite{Jhuang:ICCV:2013}, given the ground-truth position of 15 key points of human body in the first frame, our proposed method effectively tracks them in the later frames.}
    \label{fig:vis_pos}
\end{figure}

\noindent \textbf{Pose Tracking:}
We present the experimental results of pose tracking in Table \ref{tab:pose_tracking}, where our STFC$^3$ achieves the state-of-the-art of 62.0 PCK@0.1 and 80.5 PCK@0.2 under the self-supervised setting.
Compared with the most related work CRW~\cite{jabri2020walk} (the results are based on the officially released model) , our STFC$^3$ obtains 3.2 improvements on PCK@0.1, which strongly demonstrates its superiority of pose tracking task. We display some qualitative results for pose tracking in Figure \ref{fig:vis_pos}.

\begin{table}[t]
	\centering
	\caption{Face landmark tracking results on the 300VW dataset~\cite{DBLP:conf/iccvw/ShenZCKTP15}, where the lower $\downarrow$ is better.}
		\label{tab:face_landmark_tracking}
		\vspace{2mm}
	\begin{tabular}{lcc}
			\hline
{Method}  & RMSE $\downarrow$ & Supervision\\
			\hline
STRRN~\cite{DBLP:conf/aaai/ZhuLYS20} & 5.31 & 300W~\cite{DBLP:conf/iccvw/SagonasTZP13} \\
SBR~\cite{DBLP:conf/cvpr/DongYWW0S18} & 5.77 &  300W~\cite{DBLP:conf/iccvw/SagonasTZP13} + ImageNet\\
ResNet-18 & 22.8 & ImageNet\\ 
			\hline
MoCo~\cite{he2019moco} & 23.3 & self-supervision\\ 
VINCE~\cite{gordon2020watching} & 23.4 & self-supervision\\ 
			\hline
CRW~\cite{jabri2020walk} & 21.6 & self-supervision\\ 
UVC~\cite{li2019joint} & 19.9 & self-supervision\\ 
STFC$^3$ (Ours) & \textbf{18.8} & self-supervision\\ 
\hline
	\end{tabular}
\end{table}

\begin{figure*}[t]
    \centering
    \includegraphics[width=0.9\linewidth]{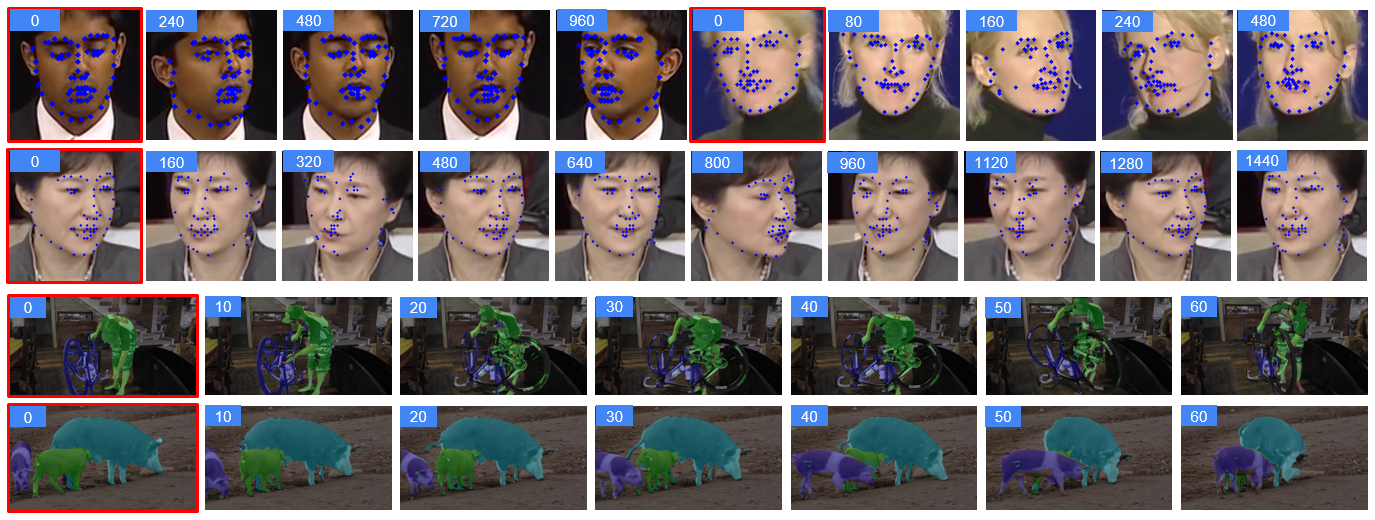}
    \caption{Qualitative results of our method on 300VW~\cite{DBLP:conf/iccvw/ShenZCKTP15} for face landmark tracking and DAVIS-2017~\cite{pont-tuset2017} for video object segmentation.
    The ground-truth labels in the first frame with a red outline are propagated to the later frames.
    }
    \vspace{-2mm}
    \label{fig:qualitative}
\end{figure*}

\begin{table}[t]
	\centering
	\caption{Evaluation on the DAVIS-2017 dataset for video object segmentation. I, C, D, K, P, M, Y represents ImageNet, COCO, DAVIS2017, Kinetics, PASCAL-VOC, Mapillary and YouTube-VOS. The methods with * are under fully-supervised learning setting. All the self-supervised methods are based on ResNet-18.}
	\vspace{2mm}
	\begin{tabular}{lcccc}
			\hline
{Method}  & {Train Data} & $\mathcal{J}${\&}$\mathcal{F}_\textrm{m}$ & $\mathcal{J}_\textrm{m}$   & $\mathcal{F}_\textrm{m}$  \\
			\hline
PReMVOS*~\cite{luiten2018premvos} & I/C/D/P/M & 77.8 & 73.9 & 81.8 \\
STM*~\cite{oh2019video} & I/D/Y & 81.8 & 79.2 & 84.3 \\
CFBI*~\cite{DBLP:conf/eccv/YangW020} & I/C/D & 83.3 & 80.5 & 86.0 \\
ResNet-18*~\cite{He2016}                                                           & I     & 62.9           & 60.6                 & 65.2            \\
			\hline
MoCo ~\cite{he2019moco}                                                          & I    & 60.8           &      58.6     &      63.1             \\
VINCE ~\cite{gordon2020watching}                                                 & K    & 60.4            & 57.9            &      62.8 \\
\hline
Colorization~\cite{vondrick2018tracking}                                        & K    & 34.0            & 34.6                & 32.7                \\
TimeCycle~\cite{wang2019learning}                                      & VLOG         & 48.7            & 46.4               & 50.0               \\
CorrFlow~\cite{lai2019self}                                        & OxUvA        & {50.3}          & {48.4}               & {52.2}                \\
UVC+track~\cite{li2019joint}                                            & K  & {59.5}          & {57.7}            & {61.3}            \\
{MAST~\cite{DBLP:conf/cvpr/LaiLX20}}                                                      & Y    & {65.5}          & {63.3}           & {67.6}            \\
VFS\cite{VFS}  & K  & {66.6} & {64.0} & {69.4} 
\\
CRW\cite{jabri2020walk}  & K  & {67.6} & {64.8} & {70.2} 
\\
STFC$^3$ (Ours)   & K & 60.5 & 58.0 & 63.1  \\
\hline
	\vspace{-5mm}
	\end{tabular}
\label{table:vos}	
\end{table}

\begin{figure*}[t]
    \centering
    \includegraphics[width=0.9\linewidth]{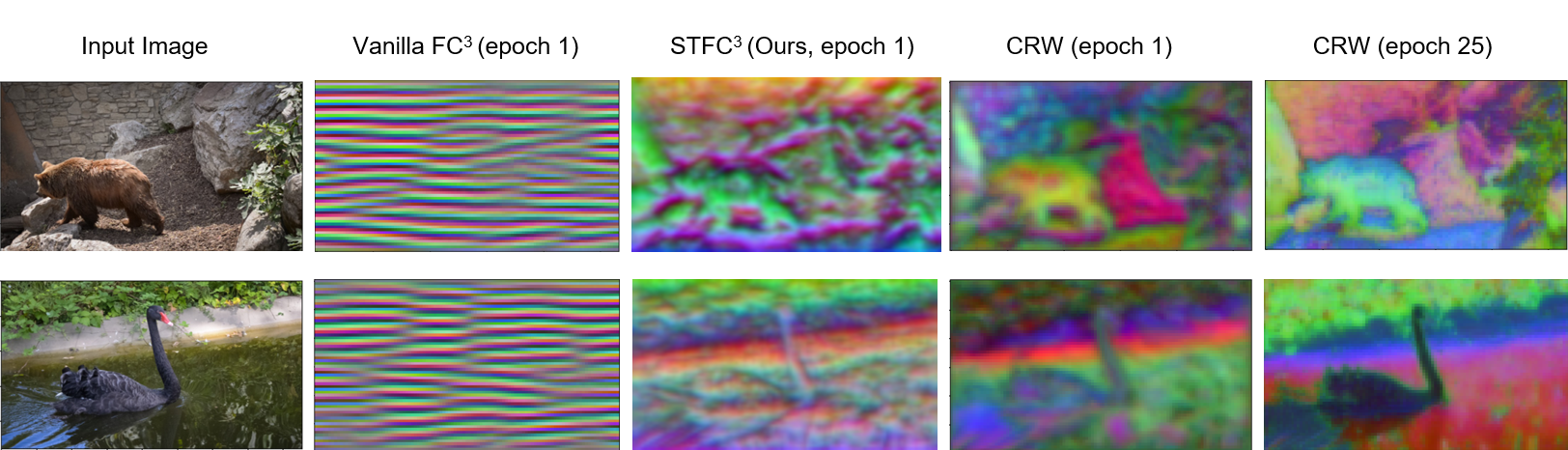}
    \caption{Visualization of the top 3 PCA components of the feature map learned by different self-supervisedly pretrained models.}
    \vspace{-2mm}
    \label{fig:visualization}
\end{figure*}

\noindent \textbf{Face Landmark Tracking:}
As shown in Table \ref{tab:face_landmark_tracking}, compared with the methods without using any supervision or initialized models pretrained on other datasets, our STFC$^3$ achieves the state-of-the-art result of 18.8 RMSE, which indicates its advantages for unsupervised face landmark tracking. However, we notice there is still a performance gap compared with the STRRN~\cite{DBLP:conf/aaai/ZhuLYS20} and SBR~\cite{DBLP:conf/cvpr/DongYWW0S18}. This is because these methods used the face detector pretrained on the 300W dataset, which introduce strong prior knowledge for this challenging task.

\noindent \textbf{Video Object Segmentation:}
We conduct experiments on the DAVIS-2017 dataset for video object segmentation (VOS). We present the quantitative results compared with the current VOS methods in Table \ref{table:vos} and some visualization results in Figure \ref{fig:qualitative}. Our proposed STFC$^3$ method achieves a performance of 60.5 $\mathcal{J}${\&}$\mathcal{F}_\textrm{m}$ with only 1 training epoch on the Kinetics dataset, while the CRW requires 25 epochs to achieve the state-of-the-art result.

While we note the CRW approach performs better than our STFC$^3$ method, it turns out that it learns invariant patch-level region features. On the other hand, our FCN based method performs pixel-level learning as shown in Figure \ref{fig:visualization} and achieves better results in keypoint tracking shown in Table \ref{tab:pose_tracking} and Table \ref{tab:face_landmark_tracking}, which is the more important aim of this paper.

\subsection{Ablation Study}
We further study different variants of our STFC$^3$ in the validation set of DAVIS-2017. Table \ref{tab:ablation} presents the experiments ablation results. We first find that using data augmentation strategy will lead to 1.3 $\mathcal{J}${\&}$\mathcal{F}_\textrm{m}$ improvement. This is consistent with the finding obtained by previous image-based self-supervised learning works~\cite{he2019moco, DBLP:conf/icml/ChenK0H20, BYOL, cao2020PIC, DBLP:conf/nips/CaronMMGBJ20}.

Last but not least, since the crop operation is an important process of our method, we conduct experiments on different arranged ratios of crop size area $\gamma$, fixing the upper bound to 1.
When the lower bound is relatively large, the model obtains poor result (\textit{e.g.,} 49.6 $\mathcal{J}${\&}$\mathcal{F}_\textrm{m}$ for $\gamma$ = [0.75, 1.0]). When the lower bound decreases, the model achieves better accuracy, and reaches the peak of 60.5 $\mathcal{J}${\&}$\mathcal{F}_\textrm{m}$ when $\gamma$ = [0.08, 1.0]. This is reasonable, because using a small lower bound of $\gamma$ results in a large variance of the crops positions, which is more beneficial for breaking the shortcut solution caused by the position encoding.
Meanwhile, when the lower bound of $\gamma$ becomes too small, the performance will slightly drop since the probability of no overlap between two crops (no correspondence) might be too high.

\begin{table}[t]
	\centering
	\caption{Ablation study on the color augmentation and the arranged ratio of cropsize area $\gamma$. All the model are evaluated on DAVIS-2017 validation set.}
	    \vspace{2mm}
	\begin{tabular}{lc}
\hline
{Method}  & $\mathcal{J}${\&}$\mathcal{F}_\textrm{m}$ \\
\hline
STFC$^3$ & 60.5\\
STFC$^3$ w/o color augmentation & 59.2\\ 
\hline
STFC$^3$, $\gamma$ = [0.75, 1.0] & 49.6\\
STFC$^3$, $\gamma$ = [0.50, 1.0] & 54.6\\
STFC$^3$, $\gamma$ = [0.25, 1.0] & 59.5\\
STFC$^3$, $\gamma$ = [0.08, 1.0] & \textbf{60.5}\\
STFC$^3$, $\gamma$ = [0.01, 1.0] & 60.1\\
			\hline
\label{tab:ablation}
\end{tabular}
\vspace{-4mm}
\end{table}

\section{Conclusion}
In this paper, we have explored using fully convolutional cycle-consistency methods for self-supervised video correspondence learning. While directly doing this leads to a shortcut solution, we have analysed the underline reason of this phenomenon, and proposed a spatial transformation approach to address this issue. Our method has largely improved the performance of the vanilla model, and achieved very competitive performance on J-HMDB dataset for pose tracking, 300VW dataset for face landmark tracking and DAVIS-2017 dataset for video object segmentation. 
This work has demonstrated the potential of using the fully convolutional method to perform cycle-consistency learning, and we hope it will shed light in this direction for future works involving more dense learning objectives.

\section{Acknowledgement}
This work is supported by the EPSRC grant/Turing AI Fellowship EP/W002981/1, EPSRC/MURIgrant EP/N019474/1, and the Tsinghua Zijing scholar Fellowship. We would like to thank the Royal Academy of Engi-neering, Tencent, and FiveAI. We would also like to thank Allan Jabri, Prof. Hao Liu, Congcong Zhu, Dr. Zheng Zhu and Xueting Li for helpful discussion. 

\begin{appendices}
\section{Qualitative Comparison to Previous State-of-the-art}
\begin{figure}[!htb]
 \begin{center}
    \includegraphics[width=1.0\linewidth]{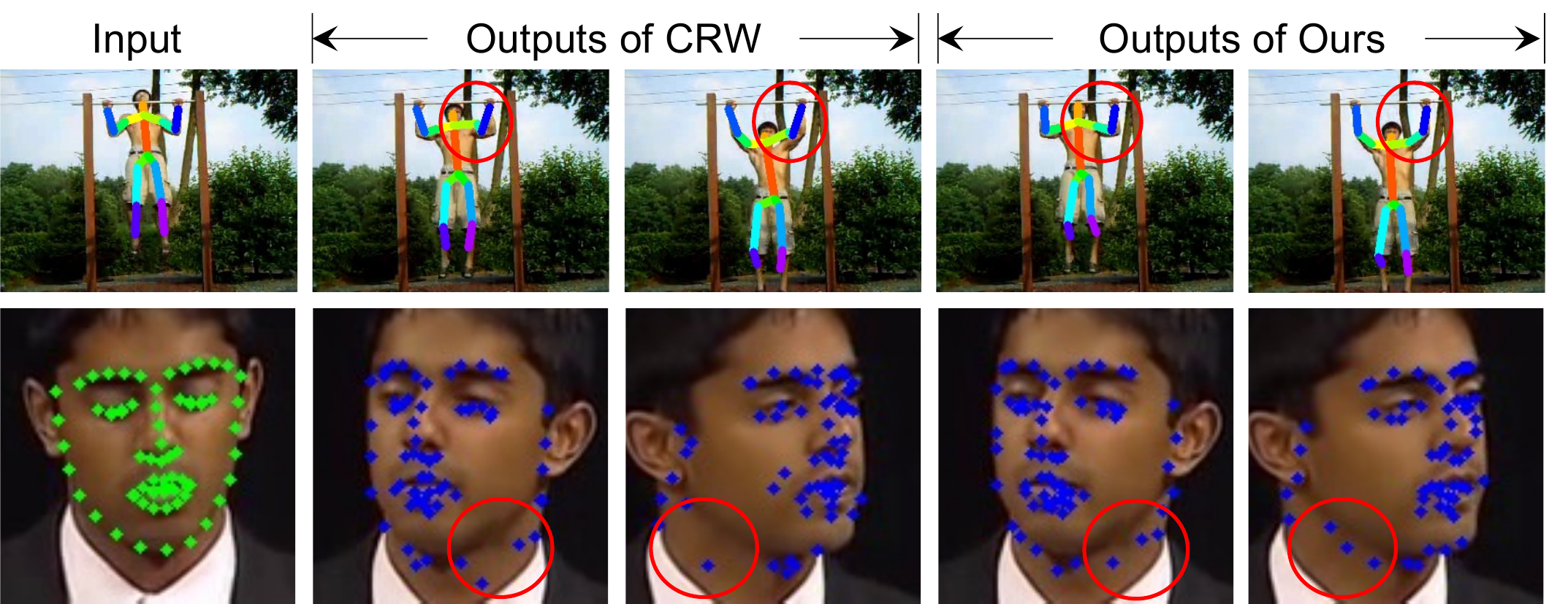}
 \end{center}    
    \caption{Visualization comparison of our method with previous state-of-the-art CRW~\cite{jabri2020walk}.}
    \label{fig:qualitative}
\end{figure} 
We show two qualitative comparison examples in Figure~\ref{fig:qualitative}. 
The two examples are about pose tracking on J-HMDB and face landmark tracking on 300VW, respectively. 
Red circle areas indicate the better localization results by our method compared with those of CRW. Our approach is more precise in localizing elbows, and also more accurate at neck area of a face. Please zoom in to have a better view. 

\section{Color Augmentations}
We follow \cite{BYOL} and use two sets of color augmentations for forward and backward frames. For forward frames, we use the following augmentations:
\begin{itemize}
\item Randomly apply color jitter with a probability of 0.8. The parameters of color jitter are brightness 0.8, contrast 0.8, saturation 0.8, hue 0.2.
\item Randomly convert image to grayscale with a probability of 0.2. The sigma of the Gaussian kernel is uniformly sampled in [0.1, 2.0].
\item Randomly apply Gaussian blur to image with a probability of 0.2.
\end{itemize}

For backward frames, we use the following augmentations:
\begin{itemize}
\item Randomly apply color jitter with a probability of 0.8. The parameters of color jitter are brightness 0.4, contrast 0.4, saturation 0.2, hue 0.1.
\item Randomly convert image to grayscale with a probability of 0.2.
\item Randomly apply Gaussian blur to image with a probability of 0.2.
\item Randomly solarize the image with a probability of 0.2.
\end{itemize}

We apply the color augmentations to each frames independently.

\section{Inference Hyper-parameters}
We provide some hyper-parameters we used during the inference stages in Table~\ref{tab:hyper}, including the context length of $m$, the spatial radius of source nodes $r$, and the number of neighours $k$. We chose them mainly following the prior work~\cite{jabri2020walk}.

\begin{table}[t]
	\centering
	\caption{Hyper-parameters used for different downstream tasks.}
	    \vspace{2mm}
	\begin{tabular}{lccc}
\hline
Task  & $m$ & $r$ & $k$ \\
\hline
Pose Tracking & 7 & 10 & 10\\
Face Landmark Tracking & 4 & 5 & 10\\
Video Object Segmentation & 20 & 18 & 10\\
			\hline
\label{tab:hyper}
	\end{tabular}
\vspace{-1.5em}
\end{table}
\end{appendices}

{\small
\bibliographystyle{ieee_fullname}
\bibliography{egbib}
}

\end{document}